%% file: main.tex
\documentclass{article} 
\usepackage{iclr2024_conference,times}

\input{math_commands.tex}

\usepackage[colorlinks]{hyperref}
\usepackage{url}
\usepackage{enumitem}		
\usepackage{graphicx}		
\usepackage{booktabs}       
\usepackage{tikz}
\usepackage{makecell}
\usepackage{graphicx}       
\usepackage{longtable}
\usetikzlibrary{shapes,arrows,positioning}
\usepackage[most]{tcolorbox}
\usepackage{xcolor}
\usepackage[framemethod=tikz]{mdframed}

\definecolor{mydarkblue}{rgb}{0,0.08,0.45}
\hypersetup{%
colorlinks=true,
linkcolor=mydarkblue,
citecolor=mydarkblue,
filecolor=mydarkblue,
urlcolor=mydarkblue}

\usepackage{amssymb}
\usepackage{pifont}

\gdef\Sepline{%
  \par\noindent\makebox[\linewidth][l]{%
  \hspace*{-\mdflength{innerleftmargin}}%
   \tikz\draw[thick,dashed,gray!60] (0,0) --%
        (\textwidth+\the\mdflength{innerleftmargin}+\the\mdflength{innerrightmargin},0);
  }\par\nobreak}

\usepackage{listings}

\definecolor{mybrown}{RGB}{128,64,0}

\newcommand*\samethanks[1][\value{footnote}]{\footnotemark[#1]}

\title{OpenWebMath: An Open Dataset of \\High-Quality Mathematical Web Text}

\author{$^{\clubsuit}$Keiran Paster\thanks{Keiran and Marco created the dataset and Zhangir led model training and evaluation.},\hspace{1em}
$^{\dagger}$Marco Dos Santos\samethanks,\hspace{1em} 
$^{\circ}$Zhangir Azerbayev, \hspace{1em}
$^{\clubsuit}$Jimmy Ba\vspace{0.2em}\\
$^{\clubsuit}$University of Toronto; Vector Institute for Artificial Intelligence\\$^{\dagger}$University of Cambridge, $^{\circ}$Princeton University\\
\texttt{keirp@cs.toronto.edu, mjad3@cam.ac.uk}
}

\iclrfinalcopy 
\begin{document}

\maketitle

\begin{abstract}

There is growing evidence that pretraining on high quality, carefully thought-out tokens such as code or mathematics plays an important role in improving the reasoning abilities of large language models. For example, Minerva, a PaLM model finetuned on billions of tokens of mathematical documents from arXiv and the web, reported dramatically improved performance on problems that require quantitative reasoning. However, because all known publicly released web datasets employ preprocessing that does not faithfully preserve mathematical notation, the benefits of large scale training on quantitive web documents are unavailable to the research community. We introduce OpenWebMath, an open dataset inspired by these works containing 14.7B tokens of mathematical webpages from Common Crawl. We describe in detail our method for extracting text and \LaTeX{} content and removing boilerplate from HTML documents, as well as our methods for quality filtering and deduplication. Additionally, we run small-scale experiments by training 1.4B parameter language models on OpenWebMath, showing that models trained on 14.7B tokens of our dataset surpass the performance of models trained on over 20x the amount of general language data. We hope that our dataset, \href{https://huggingface.co/datasets/open-web-math/open-web-math}{openly released on the Hugging Face Hub}, will help spur advances in the reasoning abilities of large language models.
\end{abstract}

\input{1_introduction}
\input{2_related_work}
\input{3_method}
\input{4_analysis}
\input{5_conclusion}
\newpage
\input{ack}
\clearpage
\bibliography{iclr2024_conference}
\bibliographystyle{iclr2024_conference}

\clearpage
\appendix

\input{6_appendix}

\end{document}

%% file: math_commands.tex
\usepackage{amsmath,amsfonts,bm}

\def\eqref#1{equation~\ref{#1}}

\def\1{\bm{1}}

\DeclareMathAlphabet{\mathsfit}{\encodingdefault}{\sfdefault}{m}{sl}
\SetMathAlphabet{\mathsfit}{bold}{\encodingdefault}{\sfdefault}{bx}{n}



%% file: 1_introduction.tex
\section{Introduction}
Advances in large language models have opened up new opportunities in numerous fields, providing a transformative shift in our approach to a wide range of complex problems \citep{gpt3, raffel2020exploring}. Among these problems, mathematical reasoning has drawn the attention of several researchers in recent years, becoming both a common benchmark to judge the performance of large language models and inspiring new approaches to improve their reasoning capabilities in the hope that they will one day be able to solve complex mathematical problems. One of the biggest advancements in mathematical reasoning in recent years has been the Minerva model \citep{lewkowycz2022solving}, which achieved state-of-the-art results on quantitative reasoning benchmarks such as MATH \citep{mathdataset}. Minerva was trained by finetuning PaLM \citep{palm} on a curated dataset consisting of billions of tokens of high quality technical content sourced from both scientific papers and the web.

Minerva and the datasets used for its training were not released publicly and the current capabilities of open-source models (e.g., \citet{touvron2023llama, llama2, codellama, openlm2023openllama, pythia}) in quantitative reasoning lags behind. We believe that there are important research directions that can only be enabled through open-access to such models and datasets, such as work on memorization and generalization, reinforcement learning, the development of new reasoning benchmarks, and advancement in the reasoning capabilities of language models.

In our work, we produce an open alternative to the Math Web Pages dataset used to train Minerva \citep{lewkowycz2022solving}. We extract documents from Common Crawl\footnote{\url{https://commoncrawl.org/}}, applying our pipeline to extract text while preserving mathematical content in the form of \LaTeX{} equations. We then filter the documents, ensuring that only high-quality English mathematical documents are kept. Finally, we deduplicate the dataset, resulting in 14.7B tokens of high-quality mathematical content suitable for both pretraining and finetuning large language models. The key contributions of this work are as follows:

\begin{itemize}
    \item We publically release OpenWebMath, a dataset of 14.7B tokens of high-quality mathematical web text. Our dataset can be found at \href{https://huggingface.co/datasets/open-web-math/open-web-math}{https://huggingface.co/datasets/open-web-math/open-web-math} on the Hugging Face Hub.
    \item We extensively document our pipeline, sharing our findings with the NLP community. We open-source the code needed to reproduce our results.
    \item We analyze the quality of OpenWebMath. First, we analyze the contents of our dataset, providing statistics on the types of webpages, subjects, and top domains. Then, we train several language models on our dataset to show that per-token, it is more effective than existing mathematical pretraining datasets, and is most effective when combined with other datasets.
\end{itemize}

%% file: 2_related_work.tex
\begin{figure}[t!]
\begin{center}
\includegraphics[width=\textwidth]{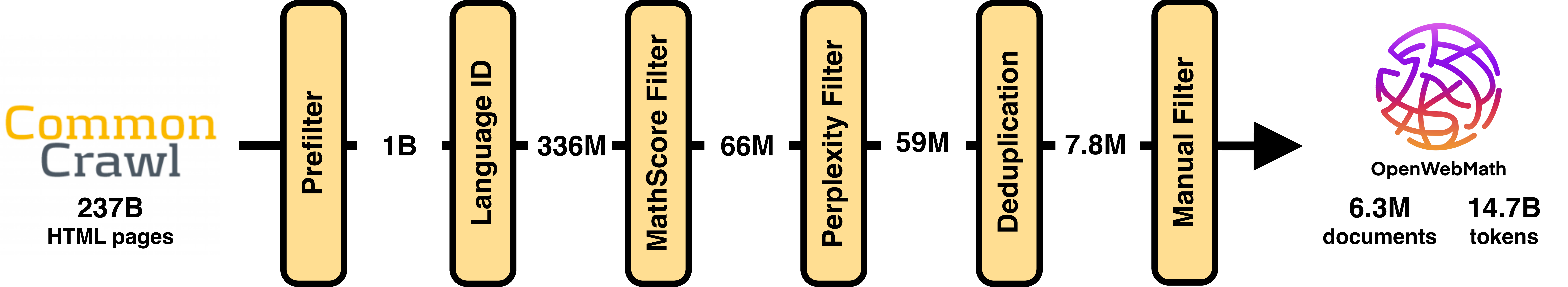}
\end{center}
\caption{The pipeline for constructing OpenWebMath involves aggressive filtering so that the final dataset only contains high quality, English, and mathematical content.}
\label{fig:pipeline}
\end{figure}

\section{Related Work}
\subsection{Mathematics datasets and benchmarks} 
\paragraph{Mathematics datasets} Over the past couple of years, several datasets of mathematics have been introduced. AMPS, a dataset of informal mathematics, was introduced alongside the MATH dataset \citep{mathdataset}. AMPS includes more than 100,000 Khan Academy problems with step-by-step solutions in LaTeX and over 5 million problems generated using Mathematica scripts. In total, AMPS contains 23GB of problems and solutions. Another notable example is NaturalProofs \citep{welleck2021naturalproofs}, which encompasses 32,000 theorem statements and proofs, 14,000 definitions, and 2,000 other types of pages (e.g. axioms, corollaries) derived from ProofWiki, the Stacks project and data from mathematics textbooks. Proof-Pile \citep{azerbayev2023proofnet} is a dataset of mathematical text that contains more than 14.5GB of informal mathematics texts obtained from arXiv, Stack Exchange, ProofWiki, Wikipedia, openly licensed books, and the MATH dataset. There are also many proprietary datasets for mathematics. WebMath is a large-scale dataset mentioned by OpenAI researchers \citep{polu2020generative} that contains a 35B token mix of content from Github, arXiv, and Math StackExchange, adding up to 35GB of informal mathematics. MathMix is another OpenAI dataset used to finetune GPT-4 \citep{verify-step-by-step} that contains 1B high quality mathematical tokens containing both natural and synthetic data. The proprietary web dataset used to train Minerva, called Math Web Pages \citep{lewkowycz2022solving}, was compiled by collecting 17.5B tokens from web pages that contain \LaTeX{} code.

\paragraph{Mathematics benchmarks} Several popular benchmarks have been used by researchers to assess the capabilities of language models on both formal and informal mathematics. The MATH dataset \citep{mathdataset} is comprised of 12,500 challenging competition problems in informal language. Each problem is also accompanied by a step-by-step informal proof. Answers are delimited by the \texttt{\textbackslash boxed} environment, allowing for easier answer verification. GSM8k \citep{cobbe2021training} is another popular multi-step informal mathematics reasoning benchmark. It contains 8,500 grade school math problems that are intended to be solvable by a bright middle school student. \citet{lewkowycz2022solving} also introduce a benchmark based on OpenCourseWare. OCWCourses includes a set of 272 automatically-verifiable solutions at the undergraduate level, covering chemistry, information theory, differential equations, special relativity, and more. \citet{lewkowycz2022solving} also evaluate on a subset of MMLU \citep{hendrycks2020measuring} called MMLU-STEM, which focuses on science, technology, engineering, and mathematics.

\begin{figure}[t!]
    \begin{center}
    \vspace{-3em}
    \includegraphics[width=\textwidth]{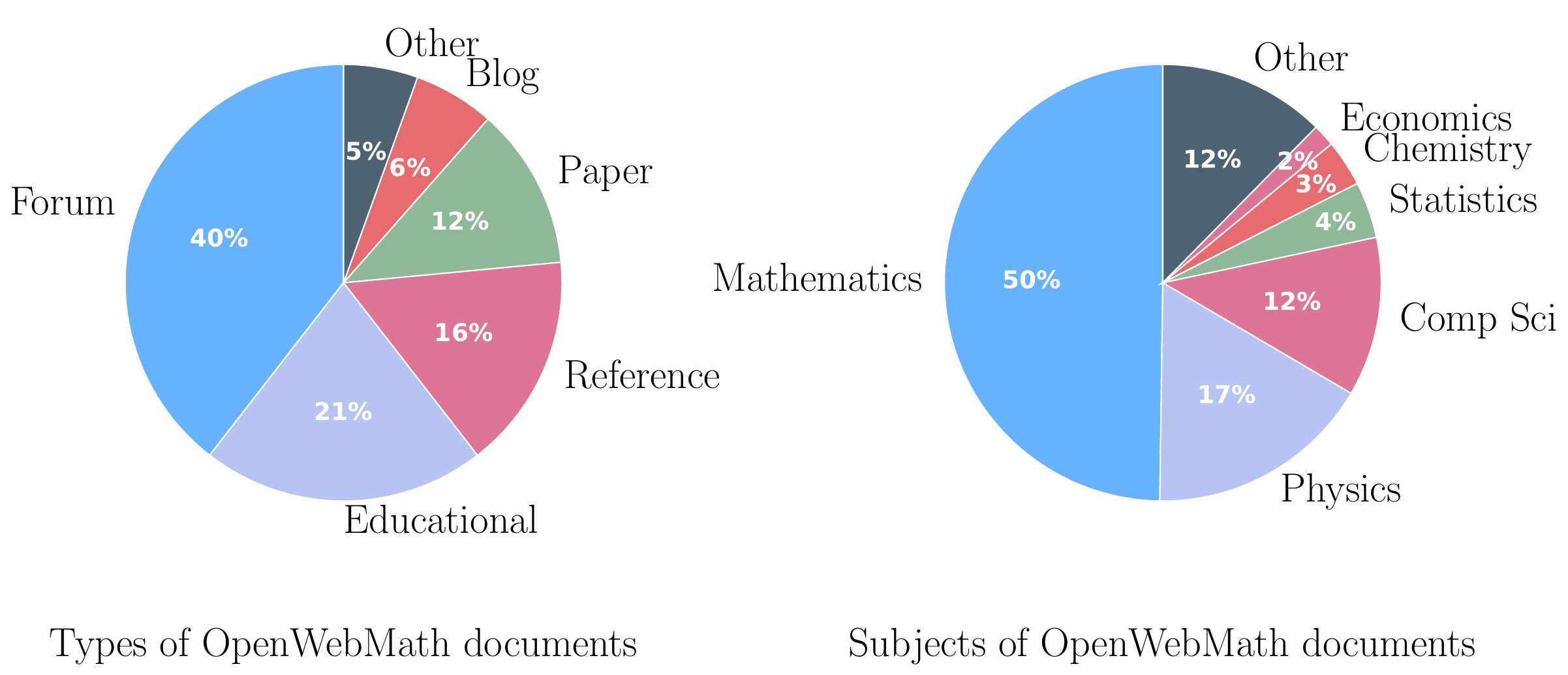}
    \end{center}
    \caption{\textbf{Left}: The documents in OpenWebMath are sourced from forum posts, educational content, reference pages, scientific papers, blogs, and more. Most content comes from Q\&A forums where users discuss how to solve problems. \textbf{Right}: The majority of the content in OpenWebMath is related to mathematics, but a large part is related to other technical subjects like Physics, Computer Science, Statistics, and more.}
    \label{fig:doublefig}
\end{figure}

\subsection{Web Data Processing Pipelines}
The pretraining of large language models requires large, diverse datasets. Data scraped from the web is one of the primary sources for such data. However, sources such as Common Crawl, which contains over 200 billion web pages, are known to have significant amounts of low-quality and duplicate content, requiring extensive filtering and deduplication to be suitable for training. Prior works such as C4 \citep{raffel2020exploring}, RefinedWeb \citep{refined-web}, CCNet \citep{wenzek2019ccnet}, The Pile \citep{gao2020pile}, and GPT-3 \citep{gpt3} introduce various pipelines for extracting quality data from Common Crawl for the purposes of language model training. These pipelines typically consist of three primary steps: text extraction, filtering, and deduplication.

\paragraph{Text extraction}
Extracting plain text from HTML files is a critical step in the creation of Common Crawl-based datasets. The easiest way to extract text from Common Crawl documents is to use the WET corresponding to each webpage, which contains pre-extracted plain text of the webpage. CCNet and C4 both use Common Crawl's WET files. However, the text extracted in WET files may contain too much boilerplate or miss out on important content such as \LaTeX{} equations. It is also possible to extract text directly from the raw HTML found in Common Crawl WARC files. The Pile uses an open source library called jusText \citep{justext} to extract text from HTML while RefinedWeb uses a library called Trafilatura \citep{barbaresi-2021-trafilatura}. These text extraction approaches differ in terms of extraction speed, customization, and their precision and recall for removing boilerplate content.

\paragraph{Filtering} 
The first layer of filtering often involves language identification \citep{wenzek2019ccnet}. Language filtering is used because certain other parts of the pipeline only work for specific languages, and is often done with simple linear classifiers such as from fastText \citep{joulin2016fasttext}. Quality filtering can be done with a combination of perplexity, classifier, and rule-based methods. CCNet uses a 5-gram Kneser-Ney language model implemented in the KenLM library \citep{heafield2011kenlm} trained on the target domain. The documents in the dataset are then sorted and filtered by their perplexity under this model. Other datasets such as the one used to train GPT-3 \citep{gpt3} use a classifier-based approach. This involves training a classifier on known-high-quality documents, such as those from Wikipedia, as positive examples and unfiltered documents from Common Crawl as negative examples. The classifier scores are used to filter low-quality documents from the dataset. Finally, rule-based approaches such as those used in C4 \citep{raffel2020exploring} and MassiveWeb \citep{gopher} involve removing pages with certain characters, too many or too few characters, too high a proportion of symbols, or those with an abnormal average word length. OpenMathWeb uses a mixture of these three approaches.

\paragraph{Deduplication} Given the periodic nature of Common Crawl snapshots and a general redundancy in web-sourced text, deduplication is an important processing step. Document-level near-deduplication (e.g., in \citep{gpt3, refined-web}) often employs MinHashLSH, an efficient algorithm for estimating the Jaccard similarity of documents. CCNet \citep{wenzek2019ccnet} uses paragraph-level deduplication, which can help to remove common boilerplate content found in WET text-extractions.

%% file: 3_method.tex
\section{Building OpenWebMath}

\subsection{Objectives}

Our aim with OpenWebMath is to build a dataset of as many mathematical documents sourced from the web as possible while preserving the formatting of mathematical content such as \LaTeX{} equations as in \citet{lewkowycz2022solving}. For the purposes of this work, we define a mathematical document as a document containing either core mathematical contents such as theorems, definitions, proofs, questions and answers, formal mathematics, or interdisciplinary documents featuring mathematical formulas within fields like physics, chemistry, biology, economics, and finance. We source our documents from Common Crawl, which is a large open-access crawl of the web containing petabytes of raw HTML files. Due to the high variance in the quality of documents from Common Crawl, we additionally use several methods for filtering and boilerplate reduction. Throughout the creation of OpenWebMath, we iteratively refined these methods to ensure that we do not remove too many relevant documents, optimizing for high recall whenever possible. Since we expect that OpenWebMath will be used primarily as an additional source of pretraining data for large language models, we prefer having a small percentage of non-mathematical but high quality documents in the dataset rather than removing them and potentially losing relevant mathematical content. Finally, due to the limited number of mathematical data available on the web, we use significantly more manual inspection and tuning of our processing pipeline than other web-based datasets. We document our processing choices and pipeline in the section that follows.

\subsection{High-level overview of the pipeline}

As shown in \autoref{fig:pipeline}, the processing pipeline for OpenWebMath falls into five stages. First, we apply a prefilter to all HTML documents in Common Crawl to quickly judge whether they have mathematical content, skipping those that do not before doing the extensive processing needed to extract text and equations and remove boilerplate. Second, we extract the text, including mathematical content, from the HTML documents. Third, we apply language identification filters, perplexity-based quality filtering, and a mathematical content classifier filter. Fourth, we deduplicate the dataset using SimHash \citep{manku2007near}. Finally, we manually inspect the documents gathered in the previous steps and view documents from the most popular domains by document-count and character-count, removing domains that are not high quality. We describe each of these steps in detail in the following sections.

\subsection{Prefiltering}

Since there are over 200B HTML documents in Common Crawl, applying our processing over each document would require a significant amount of compute. To improve the efficiency of the pipeline, we first apply a stack of pre-filters optimized for high recall to reduce the number of documents that need to be processed. Our first filters check for common mathematical strings as in \citet{lewkowycz2022solving}, such as the presence of \texttt{tex} classes, \texttt{<math>} tags, and the word ``mathjax''. See \autoref{table:math-keywords} for a full list of terms. If none of these terms are present, we search for the presence of the top 100 most-popular \LaTeX{} symbols in the text. This is done by first filtering for documents containing a backslash command using a simple regular expression and then searching specifically for these \LaTeX{} symbols in the plain text from the HTML document. If none of these symbols are found, we run the plain text through our \textit{MathScore} classifier (see \autoref{sec:math_score}) and keep documents that exceed a confidence threshold of 0.8. By tuning these filters and using hierarchical layers of progressively more accurate but more expensive filters, we were able to reduce the compute needed to process the dataset by several times while retaining a high recall of relevant documents.

\subsection{Text extraction}

In contrast with prior works that extract text from Common Crawl such as C4 \citep{collins2023evaluating}, The Pile \citep{gao2020pile}, and RefinedWeb \citep{refined-web}, we chose to make a mostly custom pipeline for extracting the main content from HTML documents. This is because we found that while other tools get decent performance on average over many documents on the internet, they do not work optimally on many of the most common sources of mathematical content on the web. We instead opted to build on top of Resiliparse \citep{bevendorff:2018, bevendorff:2021c}, a fast and efficient library built in Cython that includes performant tools for parsing HTML pages, processing their DOMs, and extracting the main content. As shown in \autoref{table:extraction-methods-comparison} in the appendix, Resiliparse is significantly more efficient than alternative libraries such as jusText. Another notable part of our text extraction pipeline is that we randomize the parameters of the extraction to add diversity to the dataset. This includes randomizing whether we use a plain text or Markdown format for the documents and randomizing the amount of boilerplate terms required to trigger a line being removed.

\input{figures/table_model_perplexity}
\input{figures/table_model_accuracy}

Our text extraction pipeline consists of four stages: \LaTeX{} extraction, text extraction, DOM processing, and line processing.

\paragraph{\LaTeX{} Extraction} \citet{lewkowycz2022solving} employ a relatively simple \LaTeX{} extraction pipeline that extracts equations from \texttt{<script type="math/latex">}, \texttt{<script type="math/asciimath">}, and \texttt{<math>} blocks with \texttt{<annotation encoding="application/x-tex">} blocks within them and replaces these tags with the extracted equations. When we applied these filters to documents from Common Crawl, we noticed an extremely low number of these tags compared to what was reported. We suspect that this is due to a difference between the HTML files available within Google \citep{lewkowycz2022solving} and those available on Common Crawl. The majority of the \LaTeX{} on the internet is written using MathJax, where developers write equations delimited by dollar signs or other delimiters in their HTML pages and then the included javascript code replaces these equations with properly rendered \LaTeX{} equations within the above script tags when the page is loaded. HTML documents on Common Crawl do not include the changes to the HTML that result from running javascript, requiring that we instead extract the \LaTeX{} equations by finding delimiters ourselves. This is a significant challenge since we need to detect whether the page contains the required MathJax javascript code, which delimiters were chosen by the user to denote equations, and then match and extract the equations from the text on the page. See Appendix \ref{appendix:text_extraction} for a more detailed discussion. 

In order to extract MathJax, we first determine whether the page is importing the MathJax javascript code by searching for the word MathJax on the page. If it is not found, we additionally search for common \LaTeX{} symbols, and if they are found, we treat the page as though it is running MathJax. We use regular expressions to search for code that calls the configuration function for MathJax to extract the delimiters used for equations. We add these delimiters to an extensive list of default delimiters and treat any content between these delimiters as \LaTeX{} equations.

In addition to extracting equations from MathJax, we found several more ways that \LaTeX{} is encoded on the internet. These methods were discovered by filtering small portions of Common Crawl for documents that contain \texttt{\textbackslash frac}, one of the most popular \LaTeX{} commands, and making sure that our processing code supports all the different ways that math could be encoded. We found that \LaTeX{} on the internet is encoded in the following ways:

\begin{figure}[t!]
\begin{center}
\includegraphics[width=\textwidth]{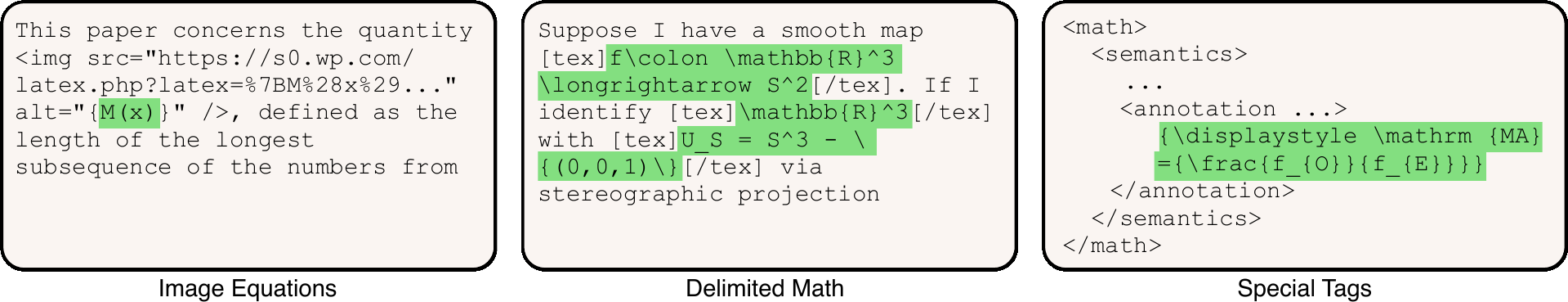}
\end{center}
\vspace{-1.2em}
\caption{\LaTeX{} formulas can be embedded in HTML documents in many ways, including in images, within arbitrary delimiters, and within special tags. Most common text-extraction pipelines do not extract \LaTeX{} code properly.}
\end{figure}

\begin{enumerate}
    \item \texttt{equation} and \texttt{align} environments.
    \item The \texttt{alttext} of elements with special classes like \texttt{tex}.
    \item Images from domains like \texttt{latex.codecogs.com} often include equations encoded in the URL.
    \item Special wordpress plugins.
    \item \texttt{<math>} tags with \texttt{<annotation encoding="application/x-tex">} blocks within them.
    \item \texttt{<math>} tags with MathML content. We use a style sheet to convert these equations into \LaTeX{}.
    \item MathJax equations encoded in the text of the page.
\end{enumerate}

The relative frequencies of the different ways math is encoded can be found in \autoref{table:latex-types} in the appendix.

\paragraph{DOM Processing} After extracting the \LaTeX{} equations from the HTML, we do several processing steps on the DOM-tree of the HTML document. This includes removing invisible elements based on their styles, removing buttons and link clusters, annotating code, tables, and headers, and removing known problematic elements based on class or ID.

\paragraph{Text Extraction} We use the \texttt{extract\_plain\_text(main\_content=True)} method in Resiliparse \citep{bevendorff:2018} to extract the main content text from the DOM following several preprocessing steps to get around common issues with their specific implementation that cause it to be overly sensitive when removing boilerplate.

\paragraph{Line Processing} After extracting the plain text on the page using Resiliparse, we apply our own processing to remove boilerplate lines based on an iteratively-refined set of common boilerplate phrases, remove empty headers, and escape dollar signs that are not part of \LaTeX{} equations.

\subsection{Filtering}
\label{sec:filtering}

We apply filtering with the goal of removing non-English documents (since our filters pipeline is optimized for English), removing documents that are not mathematical, and removing low-quality documents that would be harmful to train a language model on. We apply the following filters in order:

\begin{figure}[t!]
\begin{center}
\includegraphics[width=\textwidth]{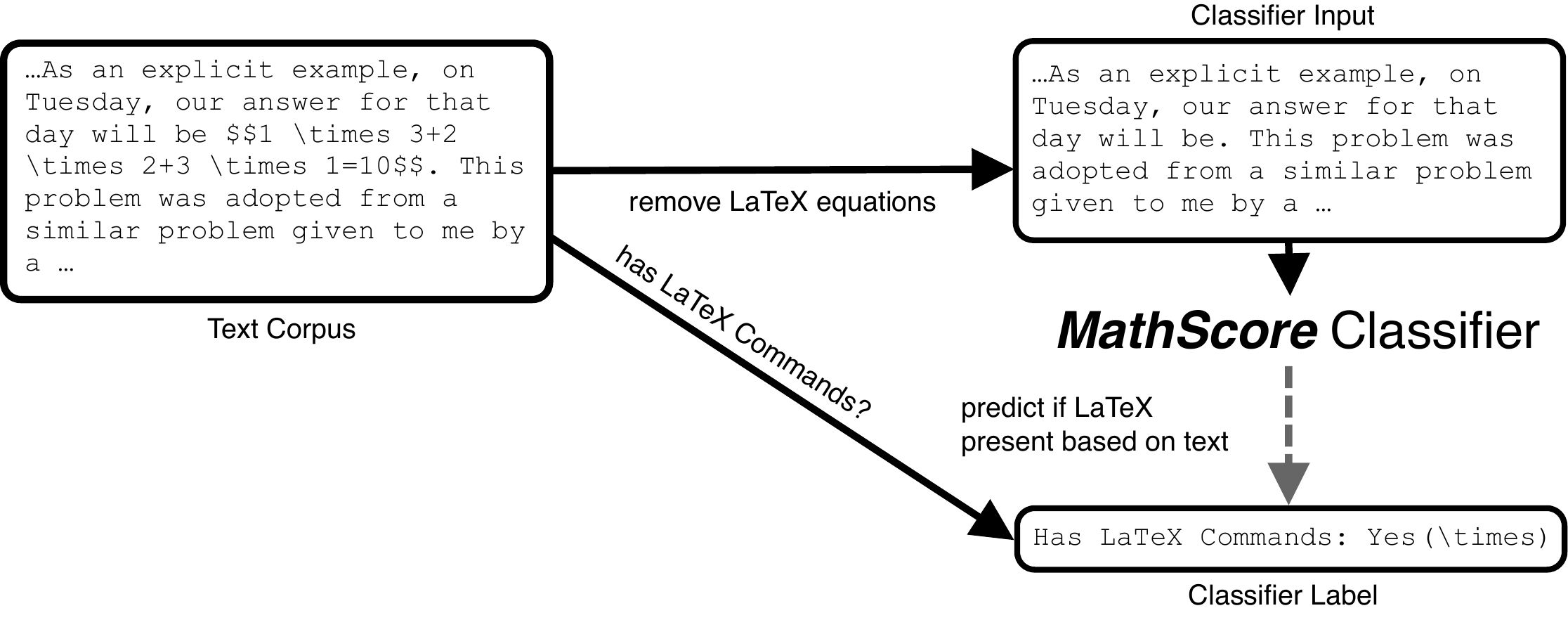}
\end{center}
\vspace{-1em}
\caption{The MathScore classifier used in filtering OpenWebMath is trained to predict whether a text has any of the most popular \LaTeX{} commands based only on surrounding words. This lets us include documents on the web that do not include extractable \LaTeX{} but still contain technical content.}
\vspace{-1em}
\label{fig:mathscore}
\end{figure}

\begin{enumerate}
    \item We use a FastText language identification model \citep{joulin2016fasttext} to remove documents that are not in English.
    \item We use our \textit{MathScore} classifier (see \autoref{sec:math_score}) to get a probability that the document is mathematical. If our previous extraction step found \LaTeX{} equations, we keep documents with a probability of over 0.17. If no \LaTeX{} equations were found, we keep documents with a probability of over 0.8.
    \item We use a KenLM language model \citep{heafield2011kenlm} trained on ProofPile \citep{azerbayev2023proofnet} to get a perplexity score for each document. We remove documents with a perplexity score of more than 15,000.
\end{enumerate}

\subsubsection{Math Score}
\label{sec:math_score}

During our filtering process, we train a model to predict the probability a document is mathematical, which we call \textit{MathScore}. We first gather a dataset of hundreds of thousands documents extracted from our pipeline from an early stage of the project, and label them depending on whether they contain one of the top-100 most common \LaTeX{} commands. We then remove any \LaTeX{} code from the documents and train a classifier to predict whether the documents contain one of these common \LaTeX{} commands. The training process for \textit{MathScore} is depicted in \autoref{fig:mathscore}. Since we remove all \LaTeX{} code from the features fed into the model, the model needs to learn the words and phrases most commonly associated with \LaTeX{} content. We use FastText \citep{joulin2016fasttext} to train this model, and find based on manual inspection that content with a score of under 0.2 is very unlikely to contain useful mathematical content.

\subsection{Deduplication}

Due to the large amount of duplicate documents in Common Crawl, we apply a deduplication step to remove near-duplicate documents. We use the SimHash implementation from text-dedup \citep{text-dedup} to deduplicate the dataset using a threshold of 0.7. We find that this threshold is high enough to remove most duplicate documents even if they have slight differences in their texts.

\subsection{Manual Inspection}

Finally, we manually inspect the top domains by document count, the top domains by character count, and the longest documents in the dataset to ensure that the documents are high quality. We remove domains that are not high quality or clearly not mathematical by adding domains to a blacklist and adding domain filters such as removing user profile pages, abstract-hosting websites as in \citet{lewkowycz2022solving}, and removing search result pages.

%% file: figures/table_model_perplexity.tex
\begin{table}[t]
\centering
\resizebox{\textwidth}{!}{
\begin{tabular}{lcccccccc}
\toprule
\textbf{Training Dataset} & \multicolumn{1}{c}{\textbf{GSM8k}} & \multicolumn{7}{c}{\textbf{MATH}} \\
                & & Prealgebra & Algebra & \makecell{Intermediate\\Algebra} & \makecell{Counting \&\\Probability} & \makecell{Number\\Theory} & \makecell{Precalculus} & Geometry\\
\midrule
\textbf{The Pile} (14.7B tokens) & 2.2032 & 1.9127 & 1.9751 & 1.8420 & 1.8193 & 1.9227 & 1.6847 & 1.9499 \\
\textbf{ProofPile} (14.7B tokens) & 2.2350 & 1.7370 & 1.7214 & 1.5739 & 1.6462 & 1.7291 & 1.4838 & 1.7229 \\
\textbf{OpenWebMath} (14.7B tokens) & 1.9075 & 1.6285 & 1.6503 & 1.5949 & 1.6002 & 1.6894 & 1.4542 & 1.5748 \\
\textbf{Mixture} (14.7B tokens) & \textbf{1.8968} & \textbf{1.6055} & \textbf{1.6190} & \textbf{1.5301} & \textbf{1.5719} & \textbf{1.6607} & \textbf{1.4119} & \textbf{1.5599} \\
\midrule
\textbf{The Pile} (300B tokens; Pythia 1.4B) & 1.9430 & 1.7117 & 1.7560 & 1.6358 & 1.6359 & 1.7460 & 1.5191 & 1.7252 \\
\bottomrule
\end{tabular}
}
\caption{We trained 1.4B parameter models for 14.7B tokens on various datasets and measured their perplexity on different mathematics benchmarks. Both OpenWebMath and a 50/50 mixture of ProofPile \cite{azerbayev2023proofnet} and OpenWebMath perform well - outperforming Pythia 1.4B \citep{pythia} trained on 300B tokens of The Pile \citep{gao2020pile}.}
\vspace{-1.5em}
\label{table:math-perplexity-comparison}
\end{table}

%% file: figures/table_model_accuracy.tex
\begin{table}[b!]
\centering
\resizebox{\textwidth}{!}{%
\begin{tabular}{l|cccc}
\toprule
\multicolumn{1}{c}{\bf Training Dataset} & \multicolumn{1}{c}{\bf MATH Algebra-Easy} & \multicolumn{1}{c}{\bf \makecell{MATH Algebra-Easy\\maj@16}} & \multicolumn{1}{c}{\bf LILA multiarith} \\
\midrule
\textbf{The Pile} (14.7B tokens) & 2.81\% & 3.93\% & 9.77\% \\
\textbf{ProofPile} (14.7B tokens) & 2.81\% & 3.93\% & 8.04\% \\
\textbf{OpenWebMath} (14.7B tokens) & \textbf{5.62\%} & 9.55\% & \textbf{16.67\%} \\
\textbf{Mixture} (14.7B tokens) & 5.06\% & \textbf{10.11\%} & 13.22\% \\
\midrule
\textbf{The Pile} (300B tokens; Pythia 1.4B) & 3.93\% & 5.62\% & \textbf{21.80\%} \\
\bottomrule
\end{tabular}
}
\caption{Accuracy on Different Math Benchmarks.}
\label{table:math-accuracy-comparison}
\end{table}

%% file: 4_analysis.tex
\section{Dataset Analysis}

\paragraph{Token count} At 14.7B tokens, OpenWebMath is just below the size of Minerva's Math Web Pages (17.5B tokens) \citet{lewkowycz2022solving} and significantly larger than the web part of any other dataset. OpenWebMath has around the same number of LLaMA tokens as ProofPile (14.2B) \citep{azerbayev2023proofnet}, but we note that there is very little overlap between between the two datasets. As a result, OpenWebMath brings a large number of new mathematical tokens that were previously unavailable to the open-source community. Due to differences in data curation strategies, it is hard to compare these datasets other than by training models on them. Since not much is known about how to properly filter a dataset, we opted to keep as much relevant content as possible. However, future work could explore filtering OpenWebMath more aggressively to further improve its quality.

\paragraph{Data Composition} We measured the distribution of domains in OpenWebMath both by document and by character count. \autoref{table:common-domains} and \autoref{table:top-domains-by-characters} show the top twenty most common domains by document and character count respectively. The most common sources of data tend to be discussion forums, blog posts, and scientific papers. We find that the distribution of characters in the dataset is distributed over 131,206 domains, with 46\% of the characters appearing in the top 100 domains.

\input{figures/contents_tables_side_by_side.tex}

In order to get a sense of the types of documents found in the dataset, we analyzed 100,000 randomly sampled documents. First, we created embeddings of this data using \texttt{all-MiniLM-L12-v2} \citep{wang2020minilm} in SentenceTransformers \citep{reimers-2019-sentence-bert}. Then, we clustered these embeddings using $k$-Means with $k=128$. Finally, we took the five closest documents to each cluster center and asked \texttt{gpt-3.5-turbo} (\href{https://platform.openai.com/docs/api-reference}{https://platform.openai.com/docs/api-reference}) to classify each cluster as Math, Physics, Statistics, Chemistry, Economics, Computer Science, or Other. We then aggregated these statistics, using the size of each cluster to get an estimate of the final number of documents in each category. We note several potential issues with this methodology, including inaccuracies stemming from using an LLM for classification, and the potential that not every document within a cluster belongs to the predicted category. \autoref{fig:doublefig} shows the results of this analysis. The majority of the documents in the dataset are directly related to mathematics, while the rest are spread out throughout physics, computer science, statistics, chemistry, and economics, with 12\% of documents not falling neatly into any of these categories.

We also used GPT to analyze the types of websites found in OpenWebMath. To do this, we took a sample of 200 documents and asked \texttt{gpt-3.5-turbo} to classify each as a Forum, Paper, Blog, Reference, Educational, Reference, or other. We also gave the document URL as a feature, since we found GPT is often able to judge the topic from the URL alone. We validated our analysis by asking GPT to do this classification on the top 100 domain names and got similar results. \autoref{fig:doublefig} shows the results. The highest proportion of documents are forum pages, where users ask and answer questions related to mathematical subjects. There is also a large proportion of educational and reference content.

\paragraph{Downstream Performance}
\label{sec:models}
We ran experiments to find out how our dataset compares to other language modeling datasets. We compare models trained on OpenWebMath for a single epoch (14.7B tokens) with models trained for the same number of tokens on The Pile \citep{gao2020pile}, a general langauge modeling dataset, and ProofPile \citep{azerbayev2023proofnet}, a dataset of both formal and informal mathematics. We also train a 50/50 mixture of ProofPile and OpenWebMath to evaluate the performance of OpenWebMath when included in a mixture of other datasets, as would be common in practice.

We train randomly initialized models with the same architecture as Pythia 1.4B \citep{pythia}. We use a batch size of 1M tokens and the same hyperparameters as Pythia otherwise. These models are evaluated on a collection of mathematics benchmarks which show signal on models of this size. This includes the subset of level-1 algebra questions from MATH, LILA-multiarith to test coding ability, and GSM8k and MATH perplexities, which scale more smoothly than accuracies. We also compare to Pythia 1.4B \citep{pythia}, which was trained on 300B tokens of The Pile \citep{gao2020pile} with the same architecture.

\autoref{table:math-perplexity-comparison} shows the results for our perplexity evaluations. There is a clear performance lead for models trained with OpenWebMath and the mixture seems to perform best. Despite Pythia being trained on over 20x the number of tokens, the performance of our models on the perplexity benchmarks far exceeds its performance, showing the potential of domain-specific models for mathematics. Similarly, \autoref{table:math-accuracy-comparison} shows the performance of the models on MATH-Algebra-Easy and LILA-multiarith \citep{lila}. OpenWebMath models outperform models that were not trained on it by a significant margin.

%% file: figures/contents_tables_side_by_side.tex
\begin{table}[b!]
    \centering
    \begin{minipage}{0.48\textwidth}
        \centering
        \resizebox{\textwidth}{!}{
        \begin{tabular}{lrr}
\multicolumn{1}{c}{\bf Domain}  &\multicolumn{1}{c}{\bf \# Documents} & \multicolumn{1}{c}{\bf \% Documents}\\ 
        \midrule
        stackexchange.com & 1,136,407 & 17.99\% \\ 
        physicsforums.com & 300,044 & 4.75\% \\ 
        mathhelpforum.com & 170,721 & 2.70\% \\ 
        socratic.org & 133,983 & 2.12\% \\ 
        mathoverflow.net & 120,755 & 1.91\% \\ 
        gradesaver.com & 96,100 & 1.52\% \\ 
        zbmath.org & 91,939 & 1.46\% \\ 
        wordpress.com & 87,876 & 1.39\% \\ 
        github.io & 81,125 & 1.28\% \\ 
        brilliant.org & 68,573 & 1.09\% \\ 
        gamedev.net & 50,560 & 0.80\% \\ 
        openstudy.com & 49,041 & 0.78\% \\ 
        gmatclub.com & 48,812 & 0.77\% \\ 
        blogspot.com & 48,036 & 0.76\% \\ 
        wikipedia.org & 46,606 & 0.74\% \\ 
        ac.uk & 41,342 & 0.65\% \\ 
        nature.com & 37,403 & 0.59\% \\ 
        aimsciences.org & 36,368 & 0.58\% \\ 
        libretexts.org & 32,216 & 0.51\% \\ 
        readthedocs.io & 31,455 & 0.50\% \\
        \end{tabular}
        }
        \caption{Most Common Domains by Document Count.}
        \label{table:common-domains}
    \end{minipage}\hfill
    \begin{minipage}{0.48\textwidth}
        \centering
        \resizebox{\textwidth}{!}{
        \begin{tabular}{lrr}
        \multicolumn{1}{c}{\bf Domain}  &\multicolumn{1}{c}{\bf \# Characters} & \multicolumn{1}{c}{\bf \% Characters}\\
                \midrule
                stackexchange.com & 4,655,132,784 & 9.55\% \\ 
                nature.com & 1,529,935,838 & 3.14\% \\ 
                wordpress.com & 1,294,166,938 & 2.66\% \\ 
                physicsforums.com & 1,160,137,919 & 2.38\% \\ 
                github.io & 725,689,722 & 1.49\% \\ 
                zbmath.org & 620,019,503 & 1.27\% \\ 
                wikipedia.org & 618,024,754 & 1.27\% \\ 
                groundai.com & 545,214,990 & 1.12\% \\ 
                blogspot.com & 520,392,333 & 1.07\% \\ 
                mathoverflow.net & 499,102,560 & 1.02\% \\ 
                gmatclub.com & 442,611,169 & 0.91\% \\ 
                gamedev.net & 426,478,461 & 0.88\% \\ 
                ac.uk & 402,111,665 & 0.83\% \\ 
                aimsciences.org & 344,716,386 & 0.71\% \\ 
                mathhelpforum.com & 319,215,756 & 0.65\% \\ 
                deepai.org & 313,512,520 & 0.64\% \\ 
                libretexts.org & 282,014,149 & 0.58\% \\ 
                readthedocs.io & 269,816,413 & 0.55\% \\ 
                tib.eu & 199,714,017 & 0.41\% \\ 
                mit.edu & 198,487,362 & 0.41\% \\
        \end{tabular}
        }
        \caption{Most Common Domains by Character Count.}
        \label{table:top-domains-by-characters}
    \end{minipage}
\end{table}

%% file: 5_conclusion.tex
\section{Conclusion}

In this paper, we describe OpenWebMath, an open dataset of 14.7B high quality mathematical documents from the web. We extensively document our pipeline, including several novel methodologies for extracting \LaTeX{} formulas, reducing boilerplate, and filtering the dataset. OpenWebMath consists of high quality Q\&A forum posts, educational documents, blogs, and more spread across mathematics, physics, computer science, and other technical domains. We also train several models on OpenWebMath and other language modeling datasets to compare the downstream performance achievable by training on our dataset. Notably, we find that models trained on OpenWebMath outperform models trained on 20x more general-domain tokens in mathematics. We hope that OpenWebMath can lead to the creation of language models with improved mathematical reasoning capabilities.

%% file: ack.tex
\section*{Acknowledgements}

JB is supported by NSERC Grant [2020-06904], CIFAR AI Chairs program, Google Research Scholar Program, and Amazon Research Award. KP is supported by an NSERC PGS-D award. Resources used in preparing this research were provided, in part, by the Province of Ontario, the Government of Canada through CIFAR, Fujitsu Limited, and companies sponsoring the Vector Institute for Artificial Intelligence (\url{www.vectorinstitute.ai/partners}). Computing resources for model training were provided by EleutherAI and Brigham Young University. We thank Finn Paster for the graphic design for the logo. We additionally thank Ziming Chen, Yuhuai Wu, Stella Biderman, Aviya Skowron, Hailey Schoelkopf, and Sean Welleck for their helpful comments.

%% file: 6_appendix.tex
\section{Limitations and Future Work}

Despite the high quality of OpenWebMath, we note several limitations and avenues for future works. First, due to the high cost of extracting data from all shards on Common Crawl, we were only able to run our pipeline once. Therefore, many of our choices are without empirical justification and we provide no ablation study. We also note that the nature of this particular type of dataset means that there are many subjective choices to be made. For instance, what counts as a mathematical document? What is a high-quality document? How do we choose the threshold for near-deduplication? For each of these, we chose several values and manually inspected a few examples to choose. Due to the cost constraints, there are also practical challenges with balancing cost with accuracy when filtering and extracting text. For instance, our prefilter reduces the number of HTML documents processed to under 1\% of the documents in Common Crawl, which may be too aggressive. We also note that OpenWebMath is an English-only dataset, which limits its applications for researchers and users who speak other languages. Finally, we note that OpenWebMath only contains the \textit{text} from math on the web, not associated figures, which can be important for solving mathematical problems \citep{openai2023gpt4}. Future work should focus on finding empirical answers to the questions of what constitutes good data, creating new, efficient filtering methodologies, and extracting images inline with math text.

\section{Text Extraction}
\label{appendix:text_extraction}

\input{figures/table_text_extract}

\paragraph{Choice of Base Text Extractor} When considering which HTML text-extraction library to use, we considered the efficiency, customization, and existing boilerplate reduction methods for each option. The most commonly used option, using WET files extracted by Common Crawl, was not an option since they do not deal with \LaTeX{} correctly and offer no customization. Other options such as jusText \citep{justext}, used in The Pile \cite{gao2020pile}, removed boilerplate too aggressively, leading to sections containing math to be discarded. Likewise, Trafilatura \citep{barbaresi-2021-trafilatura}, which was used in RefinedWeb \citep{refined-web}, had poor efficiency. We decided to go with Resiliparse \citep{bevendorff:2018} due to its balanced boilerplate removal, fast runtime, and efficient Common Crawl parsing tools. \autoref{table:extraction-methods-comparison} shows the full results for our comparison.

\input{figures/table_latex_types}

\paragraph{\LaTeX{} Extraction} \LaTeX{} code comes in many forms throughout Common Crawl HTML files. We employed an iterative process to refine our extraction rules. First, we filtered shards of Common Crawl for documents that contain the string \texttt{\textbackslash frac}. Then, we filtered those documents to find those which our extraction code found no extractable \LaTeX{}. Then, we refined our code to include additional sources of math until we were confident that we had reasonable support for all formats of \LaTeX{} in HTML documents. \autoref{table:latex-types} shows the breakdown of different common types of \LaTeX{} found in HTML documents. 

We note that most of the \LaTeX{} in OpenWebMath and across the internet is encoded using MathJax, which presents a challenge. The majority of MathJax documents use dollar sign delimiters, but most dollar signs on the web do not delimit \LaTeX{} equations. This leaves us with a few options:

\begin{itemize}
    \item Detect the use of the MathJax script in the HTML file. If the script is imported, treat dollar signs as \LaTeX{} code.
    \item Detect common \LaTeX{} commands in between dollar signs. If they are present, treat dollar signs as \LaTeX{} code.
    \item Use the \textit{MathScore} classifier to determine whether the page looks like it is talking about math. If so, treat dollar signs as \LaTeX{} code.
\end{itemize}

The first option is not always accurate since the MathJax javascript code may be nested inside of another import or named differently depending on the website. The latter two options make up for many of these cases, but can fail to detect edge cases where math equations are present but the surrounding text does not indicate that the document is mathematical. We suspect Minerva \citep{lewkowycz2022solving} gets around this issue by using HTML documents where javascript code has already been executed, in which case MathJax is converted from delimited text to explicit HTML tags that are easy to detect.

\begin{table}[t!]
    \centering
    \begin{tabular}{ccccccc}
        \toprule
        Model Size & Layers & Model Dim & Heads & Learning Rate & Batch Size \\
        \midrule
        1.4 B & 24 & 2048 & 16 & $2.0 \times 10^{-4}$ & 1M \\
        \bottomrule
    \end{tabular}
    \caption{Model Hyperparameters. We use the same architecture and hyperparameters, other than batch size, as Pythia 1.4B \citep{pythia}.}
    \label{table:model_hyperparameters}
\end{table}

\input{figures/table_math_keywords}

\section{Interplay Between Extraction and Filtering}

In prior works, we noticed many cases where suboptimal HTML text extractors were used and yet text quality remains high in the dataset. This is due to the interplay between extraction and filtering. Specifically, if a text extractor fails to extract the main text, gets the formatting wrong, or includes too much boilerplate in the extraction, then both the classification and perplexity filters can filter out such examples. This can lead to subtle biases in the dataset, where specific poorly-extracted websites are excluded entirely even though they do contain high quality content. In the case of making a mathematical dataset, failure to extract and deal with inline \LaTeX{} code properly can hurt perplexity scores and lead to these documents being filtered out. We suggest practitioners tune their text extraction pipeline on a diverse set of documents before applying filtering to avoid this bias.

\section{Model Hyperparameters}

We trained models on 14.7B tokens using the LLaMA \citep{llama2} tokenizer and the architecture described in Pythia \citep{pythia}. We train the model using the GPT-NeoX library \citep{gpt-neox-library} on 8 A100 80GB GPUs. Exact hyperparameters can be found in \autoref{table:model_hyperparameters}.

\include{datasheet}

%% file: figures/table_text_extract.tex
\begin{table}[t]
\begin{center}
\begin{tabular}{lrl}
\multicolumn{1}{c}{\bf Method}  &\multicolumn{1}{c}{\bf Runtime (s)} &\multicolumn{1}{c}{\bf Source Code Link} \\
\midrule
Resiliparse & 3.99 & \href{https://github.com/chatnoir-eu/chatnoir-resiliparse}{https://github.com/chatnoir-eu/chatnoir-resiliparse} \\
HTML-Text & 10.75 & \href{https://github.com/TeamHG-Memex/html-text}{https://github.com/TeamHG-Memex/html-text} \\
Inscripts & 19.14 & \href{https://github.com/weblyzard/inscriptis}{https://github.com/weblyzard/inscriptis} \\
BoilerPy & 24.94 & \href{https://github.com/jmriebold/BoilerPy3}{https://github.com/jmriebold/BoilerPy3} \\
jusText & 31.17 & \href{https://github.com/miso-belica/jusText}{https://github.com/miso-belica/jusText} \\
HTML2Text & 37.17 & \href{https://github.com/Alir3z4/html2text/}{https://github.com/Alir3z4/html2text/} \\
BeautifulSoup & 38.42 & \href{https://code.launchpad.net/beautifulsoup}{https://code.launchpad.net/beautifulsoup} \\
Trafilatura & 63.90 & \href{https://github.com/adbar/trafilatura}{https://github.com/adbar/trafilatura} \\
ExtractNet & 299.67 & \href{https://github.com/currentslab/extractnet}{https://github.com/currentslab/extractnet} \\
\end{tabular}
\end{center}
\caption{We measured the performance of various HTML text extraction tools on a dataset of 1k documents. Resiliparse was by far the most efficient, leading us to choose it for use in our pipeline.}
\label{table:extraction-methods-comparison}
\end{table}

%% file: figures/table_latex_types.tex
\begin{table}[t]
\begin{center}
\begin{tabular}{lr}
\multicolumn{1}{c}{\bf Math Format} & \multicolumn{1}{c}{\bf Percentage of Documents} \\
\midrule
Found at least one instance of math  & 91.42\% \\
MathJax with delimiters (inline) & 50.27\% \\
MathJax with delimiters (display) & 23.37\% \\
Math found in images & 6.96\% \\
\texttt{.math-container} & 3.94\% \\
MathML code & 3.28\% \\
\texttt{<annotation>} withing \texttt{<math>} tags & 2.35\% \\
\texttt{<mathjax>} tags & 2.24\% \\
\texttt{align} environments & 1.72\% \\
\texttt{equation} environments & 1.18\% \\
within \texttt{<script>} tags & 1.01\% \\
\texttt{alttext} property of \texttt{<math>} tags & 0.24\% \\
\end{tabular}
\end{center}
\caption{Frequencies of different types of \LaTeX{} found in OpenWebMath. The most common format of \LaTeX{} found in Common Crawl is MathJax, which uses user-defined delimiters to denote math equations. Second most common is \LaTeX{} code within either the URL or \texttt{alt} text of an \texttt{img} tag.}
\label{table:latex-types}
\end{table}

%% file: figures/table_math_keywords.tex
\begin{table}[t!]
\caption{List of Math Keywords used in the prefiltering stage.}
\label{table:math-keywords}
\begin{center}
\begin{tabular}{l}
\multicolumn{1}{c}{\bf Math Keywords} \\
\midrule
\texttt{MathJax} \\
\texttt{mathjax} \\
\texttt{<math} \\
\texttt{math-container} \\
\texttt{katex.min.css} \\
\texttt{latex.php} \\
\texttt{codecogs} \\
\texttt{tex.cgi} \\
\texttt{class="tex"} \\
\texttt{class='tex'} \\
\end{tabular}
\end{center}
\end{table}

%% file: datasheet.tex
\clearpage
\section{Datasheet}

We provide a datasheet for OpenWebMath, following the framework in \citet{gebru2021datasheets}.
\begin{longtable}{p{6cm}|p{6cm}}
    \toprule
    \multicolumn{2}{c}{\textsc{\textbf{Motivation}}} \\
    \midrule
    \textbf{For what purpose was the dataset created?} & The dataset was created to enable the training of large language models on mathematical texts, in order to improve their mathematical reasoning capabilities.\\ \midrule
    \textbf{Who created the dataset and on behalf of which entity?} & The dataset was created by the authors of this work. \\ \midrule
    \textbf{Who funded the creation of the dataset?} & Resources used in preparing this research were provided, in part, by the Province of Ontario, the Government of Canada through CIFAR, Fujitsu Limited, and companies sponsoring the Vector Institute for Artificial Intelligence (\url{www.vectorinstitute.ai/partners}). Computing resources for model training were provided by EleutherAI and Brigham Young University. \\ \midrule
    \textbf{Any other comment?} &  None. \\ \midrule
    \multicolumn{2}{c}{\textsc{\textbf{Composition}}} \\ \midrule
    \textbf{What do the instances that comprise the dataset represent?} & The instances are text documents extracted from mathematics-related webpages from Common Crawl. \\ \midrule
    \textbf{How many instances are there in total?} & In total, OpenWebMath contains 6.3 million documents. \\ \midrule
    \textbf{Does the dataset contain all possible instances or is it a sample (not necessarily random) of instances from a larger set?} & OpenWebMath doesn't contain all instances of text extracted from mathematics-related webpages from Common Crawl, as our filters can miss a non-zero proportion of such webpages. However, we expect OpenWebMath to contain most of them. \\ \midrule
    \textbf{What data does each instance consist of?} & Each instance consists of plain text and metadata including the source URL, the snapshot date, and other extraction parameters. \\ \midrule
    \textbf{Is there a label or target associated with each instance?} & No. \\ \midrule
    \textbf{Is any information missing from individual instances?} & No. \\ \midrule
    \textbf{Are relationships between individual instances made explicit?} & No. \\ \midrule
    \textbf{Are there recommended data splits?} & No. \\ \midrule
    \textbf{Are there any errors, sources of noise, or redundancies in the dataset?} & Yes, a small portion of the documents from OpenWebMath are not related to mathematics, or contain bad quality content. \\ \midrule
    \textbf{Is the dataset self-contained, or does it link to or otherwise rely on external resources?} & The dataset is entirely self-contained. \\ \midrule
    \textbf{Does the dataset contain data that might be considered confidential?} & No. \\ \midrule
    \textbf{Does the dataset contain data that, if viewed directly, might be offensive, insulting, threatening, or might otherwise cause anxiety?} & The data is filtered for quality and we do not expect that this content will be offensive, but since our filters may be imperfect we make no guarantees. \\ \midrule
    \multicolumn{2}{c}{\textsc{\textbf{Collection}}} \\ \midrule
    \textbf{How was the data associated with each instance acquired?} & The data was acquired by processing data from Common Crawl. \\ \midrule
    \textbf{What mechanisms or procedures were used to collect the data?} & We refer to the CommonCrawl website (\href{https://commoncrawl.org/}{commoncrawl.org}) for details on how they collect data. \\ \midrule
    \textbf{If the dataset is a sample from a larger set, what was the sampling strategy?} & We use all data from Common Crawl that was available before May 2023. \\ \midrule
    \textbf{Who was involved in the data collection process and how were they compensated?} & Keiran Paster and Marco Dos Santos collected the data and were compensated by their respective graduate programs. \\ \midrule
    \textbf{Over what timeframe was the data collected?} & OpenWebMath uses shards of CommonCrawl gathered between 2013 and 2023.   \\ \midrule
    \textbf{Were any ethical review processes conducted?} & No. \\ \midrule
    \multicolumn{2}{c}{\textsc{\textbf{Preprocessing}}} \\ \midrule
    \textbf{Was any preprocessing/cleaning/labeling of the data done?} & Yes. See \autoref{sec:filtering} for details. \\ \midrule
    \textbf{Was the “raw” data saved in addition to the preprocessed/cleaned/labeled data?} & Yes. \\ \midrule
    \textbf{ Is the software that was used to preprocess/clean/label the data available?} & Yes. See supplementary materials. \\ \midrule
    \multicolumn{2}{c}{\textsc{\textbf{Uses}}} \\ \midrule
    \textbf{Has the dataset been used for any tasks already?} & Yes, the data was used to train 1.4B parameter language models in \autoref{sec:models} \\ \midrule
    \textbf{Is there a repository that links to any or all papers or systems that use the dataset?} & No. \\ \midrule
    \textbf{What (other) tasks could the dataset be used for?} & We primarily envision that OpenWebMath could be useful for language model pretraining, finetuning, and evaluation. \\ \midrule
    \textbf{Is there anything about the composition of the dataset or the way it was collected and preprocessed/cleaned/labeled that might impact future uses?} & It is possible that the filtering stage of the project discarded valuable documents, such as those not written in English. This makes OpenWebMath suboptimal for creating mathematical models in other languages. \\ \midrule
    \textbf{Are there tasks for which the dataset should not be used?} & Any tasks which may considered irresponsible or harmful. \\ \midrule
    \multicolumn{2}{c}{\textsc{\textbf{Distribution}}} \\ \midrule
    \textbf{Will the dataset be distributed to third parties outside of the entity on behalf of which the dataset was created?} & Yes, the dataset will be available on the Hugging Face Hub for NLP practitioners. \\ \midrule
    \textbf{How will the dataset will be distributed?} & We will distribute the dataset on the Hugging Face Hub \\ \midrule
    \textbf{When will the dataset be distributed?} & The dataset will be available when the paper is made public. \\ \midrule
    \textbf{Will the dataset be distributed under a copyright or other intellectual property (IP) license, and/or under applicable terms of use (ToU)?} & The public extract is made available under an ODC-By 1.0 license; users should also abide to the CommonCrawl ToU: \href{https://commoncrawl.org/terms-of-use/}{https://commoncrawl.org/terms-of-use/}. \\ \midrule
    \textbf{Have any third parties imposed IP-based or other restrictions on the data associated with the instances?} & Not to our knowledge. \\ \midrule
    \textbf{Do any export controls or other regulatory restrictions apply to the dataset or to individual instances?} & Not to our knowledge. \\ \midrule
    \multicolumn{2}{c}{\textsc{\textbf{Maintenance}}} \\ \midrule
    \textbf{Who will be supporting/hosting/maintaining the dataset?} & The dataset will be hosted on the Hugging Face Hub. \\ \midrule
    \textbf{How can the owner/curator/manager of the dataset be contacted?} & \texttt{keirp@cs.toronto.edu} \\ \midrule
    \textbf{Is there an erratum?} & No. \\ \midrule
    \textbf{Will the dataset be updated?} & No. \\ \midrule
    \textbf{If others want to extend/augment/build on/contribute to the dataset, is there a mechanism for them to do so?} & No. \\ \bottomrule
    \caption{\textbf{Datasheet for OpenWebMath}, following the framework introduced by \citet{gebru2021datasheets}.}
    \label{tab:datasheet}
\end{longtable}
\clearpage

%% file: main.bbl
\begin{thebibliography}{35}
\providecommand{\natexlab}[1]{#1}
\providecommand{\url}[1]{\texttt{#1}}
\expandafter\ifx\csname urlstyle\endcsname\relax
  \providecommand{\doi}[1]{doi: #1}\else
  \providecommand{\doi}{doi: \begingroup \urlstyle{rm}\Url}\fi

\bibitem[Andonian et~al.(2023)Andonian, Anthony, Biderman, Black, Gali, Gao, Hallahan, Levy-Kramer, Leahy, Nestler, Parker, Pieler, Phang, Purohit, Schoelkopf, Stander, Songz, Tigges, Thérien, Wang, and Weinbach]{gpt-neox-library}
Alex Andonian, Quentin Anthony, Stella Biderman, Sid Black, Preetham Gali, Leo Gao, Eric Hallahan, Josh Levy-Kramer, Connor Leahy, Lucas Nestler, Kip Parker, Michael Pieler, Jason Phang, Shivanshu Purohit, Hailey Schoelkopf, Dashiell Stander, Tri Songz, Curt Tigges, Benjamin Thérien, Phil Wang, and Samuel Weinbach.
\newblock {GPT-NeoX}: Large scale autoregressive language modeling in {PyTorch}.
\newblock GitHub Repo, 9 2023.
\newblock URL \url{https://www.github.com/eleutherai/gpt-neox}.

\bibitem[Azerbayev et~al.(2023)Azerbayev, Piotrowski, Schoelkopf, Ayers, Radev, and Avigad]{azerbayev2023proofnet}
Zhangir Azerbayev, Bartosz Piotrowski, Hailey Schoelkopf, Edward~W Ayers, Dragomir Radev, and Jeremy Avigad.
\newblock Proofnet: Autoformalizing and formally proving undergraduate-level mathematics.
\newblock \emph{arXiv preprint arXiv:2302.12433}, 2023.

\bibitem[Barbaresi(2021)]{barbaresi-2021-trafilatura}
Adrien Barbaresi.
\newblock {Trafilatura: A Web Scraping Library and Command-Line Tool for Text Discovery and Extraction}.
\newblock In \emph{Proceedings of the Joint Conference of the 59th Annual Meeting of the Association for Computational Linguistics and the 11th International Joint Conference on Natural Language Processing: System Demonstrations}, pp.\  122--131. Association for Computational Linguistics, 2021.
\newblock URL \url{https://aclanthology.org/2021.acl-demo.15}.

\bibitem[Bevendorff et~al.(2018)Bevendorff, Stein, Hagen, and Potthast]{bevendorff:2018}
Janek Bevendorff, Benno Stein, Matthias Hagen, and Martin Potthast.
\newblock {Elastic ChatNoir: Search Engine for the ClueWeb and the Common Crawl}.
\newblock In Leif Azzopardi, Allan Hanbury, Gabriella Pasi, and Benjamin Piwowarski (eds.), \emph{Advances in Information Retrieval. 40th European Conference on IR Research (ECIR 2018)}, Lecture Notes in Computer Science, Berlin Heidelberg New York, March 2018. Springer.

\bibitem[Bevendorff et~al.(2021)Bevendorff, Potthast, and Stein]{bevendorff:2021c}
Janek Bevendorff, Martin Potthast, and Benno Stein.
\newblock {FastWARC: Optimizing Large-Scale Web Archive Analytics}.
\newblock In Andreas Wagner, Christian Guetl, Michael Granitzer, and Stefan Voigt (eds.), \emph{3rd International Symposium on Open Search Technology (OSSYM 2021)}. International Open Search Symposium, October 2021.

\bibitem[Biderman et~al.(2023)Biderman, Schoelkopf, Anthony, Bradley, O'Brien, Hallahan, Khan, Purohit, Prashanth, Raff, Skowron, Sutawika, and van~der Wal]{pythia}
Stella Biderman, Hailey Schoelkopf, Quentin~Gregory Anthony, Herbie Bradley, Kyle O'Brien, Eric Hallahan, Mohammad~Aflah Khan, Shivanshu Purohit, USVSN~Sai Prashanth, Edward Raff, Aviya Skowron, Lintang Sutawika, and Oskar van~der Wal.
\newblock Pythia: {A} suite for analyzing large language models across training and scaling.
\newblock In Andreas Krause, Emma Brunskill, Kyunghyun Cho, Barbara Engelhardt, Sivan Sabato, and Jonathan Scarlett (eds.), \emph{International Conference on Machine Learning, {ICML} 2023, 23-29 July 2023, Honolulu, Hawaii, {USA}}, volume 202 of \emph{Proceedings of Machine Learning Research}, pp.\  2397--2430. {PMLR}, 2023.
\newblock URL \url{https://proceedings.mlr.press/v202/biderman23a.html}.

\bibitem[Brown et~al.(2020)Brown, Mann, Ryder, Subbiah, Kaplan, Dhariwal, Neelakantan, Shyam, Sastry, Askell, Agarwal, Herbert{-}Voss, Krueger, Henighan, Child, Ramesh, Ziegler, Wu, Winter, Hesse, Chen, Sigler, Litwin, Gray, Chess, Clark, Berner, McCandlish, Radford, Sutskever, and Amodei]{gpt3}
Tom~B. Brown, Benjamin Mann, Nick Ryder, Melanie Subbiah, Jared Kaplan, Prafulla Dhariwal, Arvind Neelakantan, Pranav Shyam, Girish Sastry, Amanda Askell, Sandhini Agarwal, Ariel Herbert{-}Voss, Gretchen Krueger, Tom Henighan, Rewon Child, Aditya Ramesh, Daniel~M. Ziegler, Jeffrey Wu, Clemens Winter, Christopher Hesse, Mark Chen, Eric Sigler, Mateusz Litwin, Scott Gray, Benjamin Chess, Jack Clark, Christopher Berner, Sam McCandlish, Alec Radford, Ilya Sutskever, and Dario Amodei.
\newblock Language models are few-shot learners.
\newblock In Hugo Larochelle, Marc'Aurelio Ranzato, Raia Hadsell, Maria{-}Florina Balcan, and Hsuan{-}Tien Lin (eds.), \emph{Advances in Neural Information Processing Systems 33: Annual Conference on Neural Information Processing Systems 2020, NeurIPS 2020, December 6-12, 2020, virtual}, 2020.
\newblock URL \url{https://proceedings.neurips.cc/paper/2020/hash/1457c0d6bfcb4967418bfb8ac142f64a-Abstract.html}.

\bibitem[Chowdhery et~al.(2022)Chowdhery, Narang, Devlin, Bosma, Mishra, Roberts, Barham, Chung, Sutton, Gehrmann, Schuh, Shi, Tsvyashchenko, Maynez, Rao, Barnes, Tay, Shazeer, Prabhakaran, Reif, Du, Hutchinson, Pope, Bradbury, Austin, Isard, Gur{-}Ari, Yin, Duke, Levskaya, Ghemawat, Dev, Michalewski, Garcia, Misra, Robinson, Fedus, Zhou, Ippolito, Luan, Lim, Zoph, Spiridonov, Sepassi, Dohan, Agrawal, Omernick, Dai, Pillai, Pellat, Lewkowycz, Moreira, Child, Polozov, Lee, Zhou, Wang, Saeta, Diaz, Firat, Catasta, Wei, Meier{-}Hellstern, Eck, Dean, Petrov, and Fiedel]{palm}
Aakanksha Chowdhery, Sharan Narang, Jacob Devlin, Maarten Bosma, Gaurav Mishra, Adam Roberts, Paul Barham, Hyung~Won Chung, Charles Sutton, Sebastian Gehrmann, Parker Schuh, Kensen Shi, Sasha Tsvyashchenko, Joshua Maynez, Abhishek Rao, Parker Barnes, Yi~Tay, Noam Shazeer, Vinodkumar Prabhakaran, Emily Reif, Nan Du, Ben Hutchinson, Reiner Pope, James Bradbury, Jacob Austin, Michael Isard, Guy Gur{-}Ari, Pengcheng Yin, Toju Duke, Anselm Levskaya, Sanjay Ghemawat, Sunipa Dev, Henryk Michalewski, Xavier Garcia, Vedant Misra, Kevin Robinson, Liam Fedus, Denny Zhou, Daphne Ippolito, David Luan, Hyeontaek Lim, Barret Zoph, Alexander Spiridonov, Ryan Sepassi, David Dohan, Shivani Agrawal, Mark Omernick, Andrew~M. Dai, Thanumalayan~Sankaranarayana Pillai, Marie Pellat, Aitor Lewkowycz, Erica Moreira, Rewon Child, Oleksandr Polozov, Katherine Lee, Zongwei Zhou, Xuezhi Wang, Brennan Saeta, Mark Diaz, Orhan Firat, Michele Catasta, Jason Wei, Kathy Meier{-}Hellstern, Douglas Eck, Jeff Dean, Slav Petrov, and Noah Fiedel.
\newblock Palm: Scaling language modeling with pathways.
\newblock \emph{CoRR}, abs/2204.02311, 2022.
\newblock \doi{10.48550/arXiv.2204.02311}.
\newblock URL \url{https://doi.org/10.48550/arXiv.2204.02311}.

\bibitem[Cobbe et~al.(2021)Cobbe, Kosaraju, Bavarian, Chen, Jun, Kaiser, Plappert, Tworek, Hilton, Nakano, et~al.]{cobbe2021training}
Karl Cobbe, Vineet Kosaraju, Mohammad Bavarian, Mark Chen, Heewoo Jun, Lukasz Kaiser, Matthias Plappert, Jerry Tworek, Jacob Hilton, Reiichiro Nakano, et~al.
\newblock Training verifiers to solve math word problems.
\newblock \emph{arXiv preprint arXiv:2110.14168}, 2021.

\bibitem[Collins et~al.(2023)Collins, Jiang, Frieder, Wong, Zilka, Bhatt, Lukasiewicz, Wu, Tenenbaum, Hart, et~al.]{collins2023evaluating}
Katherine~M Collins, Albert~Q Jiang, Simon Frieder, Lionel Wong, Miri Zilka, Umang Bhatt, Thomas Lukasiewicz, Yuhuai Wu, Joshua~B Tenenbaum, William Hart, et~al.
\newblock Evaluating language models for mathematics through interactions.
\newblock \emph{arXiv preprint arXiv:2306.01694}, 2023.

\bibitem[Endrédy \& Novák(2013)Endrédy and Novák]{justext}
István Endrédy and Attila Novák.
\newblock More effective boilerplate removal-the goldminer algorithm.
\newblock \emph{Polibits}, 48:\penalty0 79--83, 12 2013.
\newblock \doi{10.17562/PB-48-10}.

\bibitem[Gao et~al.(2020)Gao, Biderman, Black, Golding, Hoppe, Foster, Phang, He, Thite, Nabeshima, et~al.]{gao2020pile}
Leo Gao, Stella Biderman, Sid Black, Laurence Golding, Travis Hoppe, Charles Foster, Jason Phang, Horace He, Anish Thite, Noa Nabeshima, et~al.
\newblock The pile: An 800gb dataset of diverse text for language modeling.
\newblock \emph{arXiv preprint arXiv:2101.00027}, 2020.

\bibitem[Gebru et~al.(2021)Gebru, Morgenstern, Vecchione, Vaughan, Wallach, au2, and Crawford]{gebru2021datasheets}
Timnit Gebru, Jamie Morgenstern, Briana Vecchione, Jennifer~Wortman Vaughan, Hanna Wallach, Hal Daumé~III au2, and Kate Crawford.
\newblock Datasheets for datasets, 2021.

\bibitem[Geng \& Liu(2023)Geng and Liu]{openlm2023openllama}
Xinyang Geng and Hao Liu.
\newblock Openllama: An open reproduction of llama, May 2023.
\newblock URL \url{https://github.com/openlm-research/open_llama}.

\bibitem[Heafield(2011)]{heafield2011kenlm}
Kenneth Heafield.
\newblock Kenlm: Faster and smaller language model queries.
\newblock In \emph{Proceedings of the sixth workshop on statistical machine translation}, pp.\  187--197, 2011.

\bibitem[Hendrycks et~al.(2020)Hendrycks, Burns, Basart, Zou, Mazeika, Song, and Steinhardt]{hendrycks2020measuring}
Dan Hendrycks, Collin Burns, Steven Basart, Andy Zou, Mantas Mazeika, Dawn Song, and Jacob Steinhardt.
\newblock Measuring massive multitask language understanding.
\newblock \emph{arXiv preprint arXiv:2009.03300}, 2020.

\bibitem[Hendrycks et~al.(2021)Hendrycks, Burns, Kadavath, Arora, Basart, Tang, Song, and Steinhardt]{mathdataset}
Dan Hendrycks, Collin Burns, Saurav Kadavath, Akul Arora, Steven Basart, Eric Tang, Dawn Song, and Jacob Steinhardt.
\newblock Measuring mathematical problem solving with the {MATH} dataset.
\newblock \emph{CoRR}, abs/2103.03874, 2021.
\newblock URL \url{https://arxiv.org/abs/2103.03874}.

\bibitem[Joulin et~al.(2016)Joulin, Grave, Bojanowski, Douze, J{\'e}gou, and Mikolov]{joulin2016fasttext}
Armand Joulin, Edouard Grave, Piotr Bojanowski, Matthijs Douze, H{\'e}rve J{\'e}gou, and Tomas Mikolov.
\newblock Fasttext.zip: Compressing text classification models.
\newblock \emph{arXiv preprint arXiv:1612.03651}, 2016.

\bibitem[Lewkowycz et~al.(2022)Lewkowycz, Andreassen, Dohan, Dyer, Michalewski, Ramasesh, Slone, Anil, Schlag, Gutman-Solo, et~al.]{lewkowycz2022solving}
Aitor Lewkowycz, Anders Andreassen, David Dohan, Ethan Dyer, Henryk Michalewski, Vinay Ramasesh, Ambrose Slone, Cem Anil, Imanol Schlag, Theo Gutman-Solo, et~al.
\newblock Solving quantitative reasoning problems with language models.
\newblock \emph{Advances in Neural Information Processing Systems}, 35:\penalty0 3843--3857, 2022.

\bibitem[Lightman et~al.(2023)Lightman, Kosaraju, Burda, Edwards, Baker, Lee, Leike, Schulman, Sutskever, and Cobbe]{verify-step-by-step}
Hunter Lightman, Vineet Kosaraju, Yura Burda, Harrison Edwards, Bowen Baker, Teddy Lee, Jan Leike, John Schulman, Ilya Sutskever, and Karl Cobbe.
\newblock Let's verify step by step.
\newblock \emph{CoRR}, abs/2305.20050, 2023.
\newblock \doi{10.48550/arXiv.2305.20050}.
\newblock URL \url{https://doi.org/10.48550/arXiv.2305.20050}.

\bibitem[Manku et~al.(2007)Manku, Jain, and Das~Sarma]{manku2007near}
Gurmeet~Singh Manku, Arvind Jain, and Anish Das~Sarma.
\newblock Detecting near-duplicates for web crawling.
\newblock In \emph{Proceedings of the 16th International Conference on World Wide Web}, WWW '07, pp.\  141–150, New York, NY, USA, 2007. Association for Computing Machinery.
\newblock ISBN 9781595936547.
\newblock \doi{10.1145/1242572.1242592}.
\newblock URL \url{https://doi.org/10.1145/1242572.1242592}.

\bibitem[Mishra et~al.(2022)Mishra, Finlayson, Lu, Tang, Welleck, Baral, Rajpurohit, Tafjord, Sabharwal, Clark, et~al.]{lila}
Swaroop Mishra, Matthew Finlayson, Pan Lu, Leonard Tang, Sean Welleck, Chitta Baral, Tanmay Rajpurohit, Oyvind Tafjord, Ashish Sabharwal, Peter Clark, et~al.
\newblock Lila: A unified benchmark for mathematical reasoning.
\newblock \emph{arXiv preprint arXiv:2210.17517}, 2022.

\bibitem[Mou et~al.(2023)Mou, Ha, Enevoldsen, and Liu]{text-dedup}
Chenghao Mou, Chris Ha, Kenneth Enevoldsen, and Peiyuan Liu.
\newblock Chenghaomou/text-dedup: Reference snapshot, September 2023.
\newblock URL \url{https://doi.org/10.5281/zenodo.8364980}.

\bibitem[OpenAI(2023)]{openai2023gpt4}
OpenAI.
\newblock Gpt-4 technical report, 2023.

\bibitem[Penedo et~al.(2023)Penedo, Malartic, Hesslow, Cojocaru, Cappelli, Alobeidli, Pannier, Almazrouei, and Launay]{refined-web}
Guilherme Penedo, Quentin Malartic, Daniel Hesslow, Ruxandra Cojocaru, Alessandro Cappelli, Hamza Alobeidli, Baptiste Pannier, Ebtesam Almazrouei, and Julien Launay.
\newblock The refinedweb dataset for falcon {LLM:} outperforming curated corpora with web data, and web data only.
\newblock \emph{CoRR}, abs/2306.01116, 2023.
\newblock \doi{10.48550/arXiv.2306.01116}.
\newblock URL \url{https://doi.org/10.48550/arXiv.2306.01116}.

\bibitem[Polu \& Sutskever(2020)Polu and Sutskever]{polu2020generative}
Stanislas Polu and Ilya Sutskever.
\newblock Generative language modeling for automated theorem proving.
\newblock \emph{arXiv preprint arXiv:2009.03393}, 2020.

\bibitem[Rae et~al.(2021)Rae, Borgeaud, Cai, Millican, Hoffmann, Song, Aslanides, Henderson, Ring, Young, Rutherford, Hennigan, Menick, Cassirer, Powell, van~den Driessche, Hendricks, Rauh, Huang, Glaese, Welbl, Dathathri, Huang, Uesato, Mellor, Higgins, Creswell, McAleese, Wu, Elsen, Jayakumar, Buchatskaya, Budden, Sutherland, Simonyan, Paganini, Sifre, Martens, Li, Kuncoro, Nematzadeh, Gribovskaya, Donato, Lazaridou, Mensch, Lespiau, Tsimpoukelli, Grigorev, Fritz, Sottiaux, Pajarskas, Pohlen, Gong, Toyama, de~Masson~d'Autume, Li, Terzi, Mikulik, Babuschkin, Clark, de~Las~Casas, Guy, Jones, Bradbury, Johnson, Hechtman, Weidinger, Gabriel, Isaac, Lockhart, Osindero, Rimell, Dyer, Vinyals, Ayoub, Stanway, Bennett, Hassabis, Kavukcuoglu, and Irving]{gopher}
Jack~W. Rae, Sebastian Borgeaud, Trevor Cai, Katie Millican, Jordan Hoffmann, H.~Francis Song, John Aslanides, Sarah Henderson, Roman Ring, Susannah Young, Eliza Rutherford, Tom Hennigan, Jacob Menick, Albin Cassirer, Richard Powell, George van~den Driessche, Lisa~Anne Hendricks, Maribeth Rauh, Po{-}Sen Huang, Amelia Glaese, Johannes Welbl, Sumanth Dathathri, Saffron Huang, Jonathan Uesato, John Mellor, Irina Higgins, Antonia Creswell, Nat McAleese, Amy Wu, Erich Elsen, Siddhant~M. Jayakumar, Elena Buchatskaya, David Budden, Esme Sutherland, Karen Simonyan, Michela Paganini, Laurent Sifre, Lena Martens, Xiang~Lorraine Li, Adhiguna Kuncoro, Aida Nematzadeh, Elena Gribovskaya, Domenic Donato, Angeliki Lazaridou, Arthur Mensch, Jean{-}Baptiste Lespiau, Maria Tsimpoukelli, Nikolai Grigorev, Doug Fritz, Thibault Sottiaux, Mantas Pajarskas, Toby Pohlen, Zhitao Gong, Daniel Toyama, Cyprien de~Masson~d'Autume, Yujia Li, Tayfun Terzi, Vladimir Mikulik, Igor Babuschkin, Aidan Clark, Diego de~Las~Casas, Aurelia Guy,
  Chris Jones, James Bradbury, Matthew~J. Johnson, Blake~A. Hechtman, Laura Weidinger, Iason Gabriel, William Isaac, Edward Lockhart, Simon Osindero, Laura Rimell, Chris Dyer, Oriol Vinyals, Kareem Ayoub, Jeff Stanway, Lorrayne Bennett, Demis Hassabis, Koray Kavukcuoglu, and Geoffrey Irving.
\newblock Scaling language models: Methods, analysis {\&} insights from training gopher.
\newblock \emph{CoRR}, abs/2112.11446, 2021.
\newblock URL \url{https://arxiv.org/abs/2112.11446}.

\bibitem[Raffel et~al.(2020)Raffel, Shazeer, Roberts, Lee, Narang, Matena, Zhou, Li, and Liu]{raffel2020exploring}
Colin Raffel, Noam Shazeer, Adam Roberts, Katherine Lee, Sharan Narang, Michael Matena, Yanqi Zhou, Wei Li, and Peter~J Liu.
\newblock Exploring the limits of transfer learning with a unified text-to-text transformer.
\newblock \emph{The Journal of Machine Learning Research}, 21\penalty0 (1):\penalty0 5485--5551, 2020.

\bibitem[Reimers \& Gurevych(2019)Reimers and Gurevych]{reimers-2019-sentence-bert}
Nils Reimers and Iryna Gurevych.
\newblock Sentence-bert: Sentence embeddings using siamese bert-networks.
\newblock In \emph{Proceedings of the 2019 Conference on Empirical Methods in Natural Language Processing}. Association for Computational Linguistics, 11 2019.
\newblock URL \url{https://arxiv.org/abs/1908.10084}.

\bibitem[Touvron et~al.(2023{\natexlab{a}})Touvron, Lavril, Izacard, Martinet, Lachaux, Lacroix, Rozi{\`{e}}re, Goyal, Hambro, Azhar, Rodriguez, Joulin, Grave, and Lample]{codellama}
Hugo Touvron, Thibaut Lavril, Gautier Izacard, Xavier Martinet, Marie{-}Anne Lachaux, Timoth{\'{e}}e Lacroix, Baptiste Rozi{\`{e}}re, Naman Goyal, Eric Hambro, Faisal Azhar, Aur{\'{e}}lien Rodriguez, Armand Joulin, Edouard Grave, and Guillaume Lample.
\newblock Llama: Open and efficient foundation language models.
\newblock \emph{CoRR}, abs/2302.13971, 2023{\natexlab{a}}.
\newblock \doi{10.48550/arXiv.2302.13971}.
\newblock URL \url{https://doi.org/10.48550/arXiv.2302.13971}.

\bibitem[Touvron et~al.(2023{\natexlab{b}})Touvron, Lavril, Izacard, Martinet, Lachaux, Lacroix, Rozi{\`e}re, Goyal, Hambro, Azhar, et~al.]{touvron2023llama}
Hugo Touvron, Thibaut Lavril, Gautier Izacard, Xavier Martinet, Marie-Anne Lachaux, Timoth{\'e}e Lacroix, Baptiste Rozi{\`e}re, Naman Goyal, Eric Hambro, Faisal Azhar, et~al.
\newblock Llama: Open and efficient foundation language models.
\newblock \emph{arXiv preprint arXiv:2302.13971}, 2023{\natexlab{b}}.

\bibitem[Touvron et~al.(2023{\natexlab{c}})Touvron, Martin, Stone, Albert, Almahairi, Babaei, Bashlykov, Batra, Bhargava, Bhosale, Bikel, Blecher, Canton{-}Ferrer, Chen, Cucurull, Esiobu, Fernandes, Fu, Fu, Fuller, Gao, Goswami, Goyal, Hartshorn, Hosseini, Hou, Inan, Kardas, Kerkez, Khabsa, Kloumann, Korenev, Koura, Lachaux, Lavril, Lee, Liskovich, Lu, Mao, Martinet, Mihaylov, Mishra, Molybog, Nie, Poulton, Reizenstein, Rungta, Saladi, Schelten, Silva, Smith, Subramanian, Tan, Tang, Taylor, Williams, Kuan, Xu, Yan, Zarov, Zhang, Fan, Kambadur, Narang, Rodriguez, Stojnic, Edunov, and Scialom]{llama2}
Hugo Touvron, Louis Martin, Kevin Stone, Peter Albert, Amjad Almahairi, Yasmine Babaei, Nikolay Bashlykov, Soumya Batra, Prajjwal Bhargava, Shruti Bhosale, Dan Bikel, Lukas Blecher, Cristian Canton{-}Ferrer, Moya Chen, Guillem Cucurull, David Esiobu, Jude Fernandes, Jeremy Fu, Wenyin Fu, Brian Fuller, Cynthia Gao, Vedanuj Goswami, Naman Goyal, Anthony Hartshorn, Saghar Hosseini, Rui Hou, Hakan Inan, Marcin Kardas, Viktor Kerkez, Madian Khabsa, Isabel Kloumann, Artem Korenev, Punit~Singh Koura, Marie{-}Anne Lachaux, Thibaut Lavril, Jenya Lee, Diana Liskovich, Yinghai Lu, Yuning Mao, Xavier Martinet, Todor Mihaylov, Pushkar Mishra, Igor Molybog, Yixin Nie, Andrew Poulton, Jeremy Reizenstein, Rashi Rungta, Kalyan Saladi, Alan Schelten, Ruan Silva, Eric~Michael Smith, Ranjan Subramanian, Xiaoqing~Ellen Tan, Binh Tang, Ross Taylor, Adina Williams, Jian~Xiang Kuan, Puxin Xu, Zheng Yan, Iliyan Zarov, Yuchen Zhang, Angela Fan, Melanie Kambadur, Sharan Narang, Aur{\'{e}}lien Rodriguez, Robert Stojnic, Sergey Edunov,
  and Thomas Scialom.
\newblock Llama 2: Open foundation and fine-tuned chat models.
\newblock \emph{CoRR}, abs/2307.09288, 2023{\natexlab{c}}.
\newblock \doi{10.48550/arXiv.2307.09288}.
\newblock URL \url{https://doi.org/10.48550/arXiv.2307.09288}.

\bibitem[Wang et~al.(2020)Wang, Wei, Dong, Bao, Yang, and Zhou]{wang2020minilm}
Wenhui Wang, Furu Wei, Li~Dong, Hangbo Bao, Nan Yang, and Ming Zhou.
\newblock Minilm: Deep self-attention distillation for task-agnostic compression of pre-trained transformers, 2020.

\bibitem[Welleck et~al.(2021)Welleck, Liu, Bras, Hajishirzi, Choi, and Cho]{welleck2021naturalproofs}
Sean Welleck, Jiacheng Liu, Ronan~Le Bras, Hannaneh Hajishirzi, Yejin Choi, and Kyunghyun Cho.
\newblock Naturalproofs: Mathematical theorem proving in natural language.
\newblock \emph{arXiv preprint arXiv:2104.01112}, 2021.

\bibitem[Wenzek et~al.(2019)Wenzek, Lachaux, Conneau, Chaudhary, Guzm{\'a}n, Joulin, and Grave]{wenzek2019ccnet}
Guillaume Wenzek, Marie-Anne Lachaux, Alexis Conneau, Vishrav Chaudhary, Francisco Guzm{\'a}n, Armand Joulin, and Edouard Grave.
\newblock Ccnet: Extracting high quality monolingual datasets from web crawl data.
\newblock \emph{arXiv preprint arXiv:1911.00359}, 2019.

\end{thebibliography}
